\documentclass[sigconf]{acmart}

\usepackage{float}

\usepackage{tabularx}

\usepackage{subcaption}

\usepackage[scaled=0.83]{beramono}

\usepackage{booktabs}
\usepackage[ruled,vlined]{algorithm2e}

\acmConference[]{}{}{}

\begin{document}

\title{Handling tree-structured text: parsing directory pages}

\author{Sarang Shrivastava}
\email{sarang.shrivastava@ny.email.gs.com}
\affiliation{ \institution{Goldman Sachs}}

\author{Afreen Shaikh}
\email{afreen.shaikh@ny.email.gs.com}
\affiliation{\institution{Goldman Sachs}}

\author{Shivani Shrivastava}
\email{shivani.shrivastava@ny.email.gs.com}
\affiliation{\institution{Goldman Sachs}}

\author{Chung Ming Ho}
\email{sam.ho@ny.email.gs.com}
\affiliation{ \institution{Goldman Sachs}}
  
\author{Pradeep Reddy}
\email{pradeep.reddy@ny.email.gs.com}
\affiliation{\institution{Goldman Sachs}}

\author{Vijay Saraswat}
\email{vijay.saraswat@ny.email.gs.com}
\affiliation{ \institution{Goldman Sachs}}

\renewcommand{\shortauthors}{Sarang, et al.}

\begin{abstract}
The determination of the {\em reading sequence} of text is fundamental to document understanding. This problem is easily solved in pages where the text is organized into a sequence of lines and vertical alignment runs the height of the page (producing multiple columns which can be read from left to right). We present a situation -- the directory page parsing problem -- where information is presented on the page in an irregular, visually-organized, two-dimensional format. Directory pages are fairly common in financial prospectuses and carry information about organizations, their addresses and relationships that is key to business tasks in client onboarding. Interestingly, directory pages sometimes have hierarchical structure, motivating the need to generalize  the {\em reading sequence} to  a {\em reading tree}.  We present solutions to the problem of identifying directory pages and constructing the reading tree, using (learnt) classifiers for text segments and a bottom-up (right to left, bottom-to-top) traversal of segments. The solution is a key part of a production service supporting automatic extraction of organization, address and relationship information from client onboarding documents.

\end{abstract}





\maketitle

\section{Introduction}

The Financial industry runs on documents. Documents are the primary carriers of information and contract across legal, trading, investing, lending, compliance, client onboarding and other functions.

The client onboarding function, in particular, requires the processing of many diverse types of documents, such as limited liability agreements, partnership agreements, prospectuses, investment management agreements. These documents are typically scanned (sometimes programmatically generated) pdfs, ranging in size from a few pages to hundreds of pages. They are put forward by a prospective client (e.g. counter-party wishing to trade with the financial institution on behalf of a {\em beneficial owner}) to establish their identity (including name and addresses), their business role (e.g. {\em adviser}, {\em partner}, {\em investment manager}, typically in relation to the beneficial owner), their {\em authority} (deriving from the beneficial owner), and their {\em capacity} to engage in the given business function(s). Till recently, the work of processing these documents and establishing the relevant information (typically in a downstream database) was done manually. But this has some obvious defects: the process is slow, expensive, of variable quality, and hard to scale (up and down), in response to changing demand. 

The {\em Cobi} service at Goldman Sachs -- in production for several months -- addresses the client onboarding document processing problem by leveraging a variety of machine learning / natural language understanding techniques. The purpose of this paper is to describe a key problem -- the {\em directory page parsing problem} -- and the solution we have designed, implemented and fielded in production. 

\subsection{The problem}
Consider {\em prospectus documents}.  {\em Prospectuses } are disclosure documents that describe financial securities on offer for potential buyers. They are published when a company makes an initial public offering, or offers new kinds of financial products (e.g. shares of a specific fund). The prospectus will typically detail the fund's objective, risk performance, distribution policy, investment strategy, and list its executive, management and legal teams. For the client onboarding task it is necessary, in particular, to extract the list of organizations provided in the prospectus, together with their address and associated {\em role} (e.g. Investment Manager, Legal Counsel, Prime Broker, Director, Administrator).

\begin{figure}[h!]
\centering
\begin{subfigure}{.5\textwidth}
  \centering
  \includegraphics[width=\linewidth]{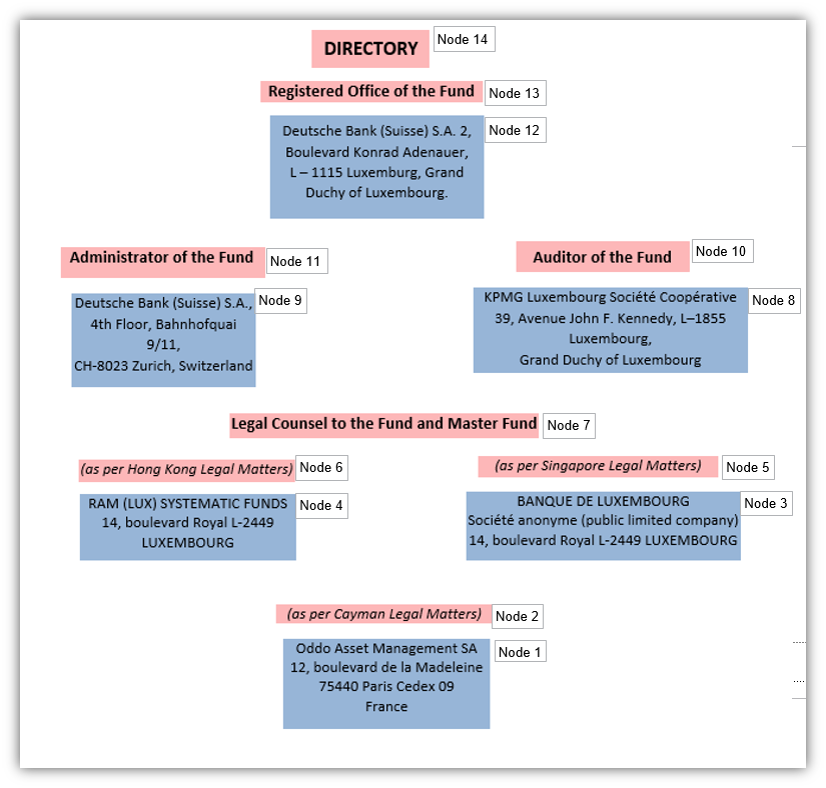}  
  \caption{Dir Page Example 1}
  \label{figure 1a}
\end{subfigure} %
\begin{subfigure}{.5\textwidth}
  \centering
  \includegraphics[width=\linewidth, height=2in]{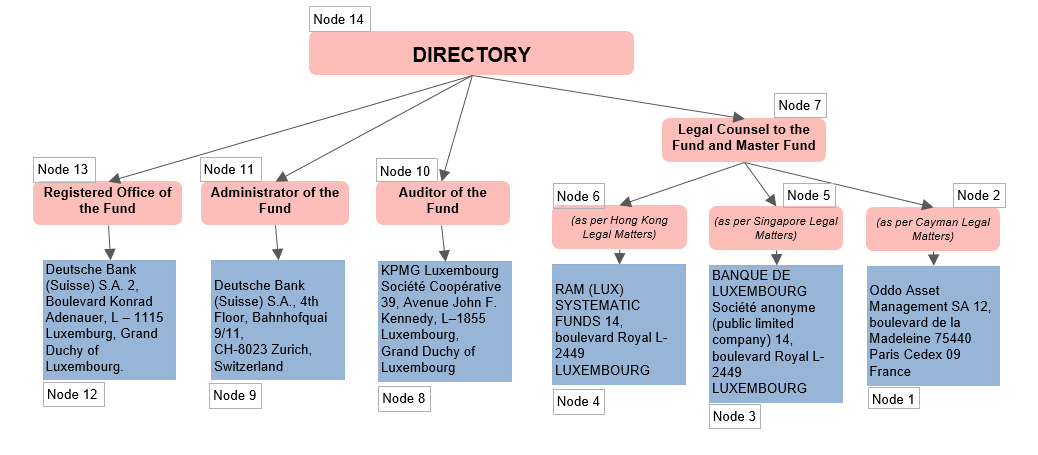}  
  \caption{Directory Tree for Example 1}
  \label{figure 1b}
\end{subfigure}
\caption[Caption]{Directory Page and its Reading Tree\footnotemark[1]}
\label{figure 1}
\end{figure}

  This information may be present in the document in a variety of ways. First some roles may be specified as part of narrative text. While in some cases the organization and role appear in close proximity (see Figure~\ref{figure 1a}), in others there may be large intervening parenthetical remarks that must be accounted for. In other cases, the information is presented through a table, listing organization, role and address in rows, typically with easily identifiable keys.

  By far the biggest challenge is offered by {\em directory pages}, illustrated in Figure~\ref{figure 1a}. Here the author chooses to provide the information hierarchically, using 2-d visual information to carry some of the structure. One can look at such a page as consisting of multiple {\em entries} with an entry specifying (typically) an organization and an address and a {\em header} which provides some context for the entry, typically a portion of the role played by the organization (e.g.{} ``Administrator of the Fund'').  Notice that portions of the page may be marked off, typically with a center panel running across the columns (e.g.{} ``Legal Counsel to the Fund and Master Fund''); one can think of this as an entry with only a header.  The role associated with an organization must sometimes be obtained by combining headers from multiple entries (e.g.{} ``Legal Counsel to the Fund and Master Fund'' / ``(as per Hong Kong legal matters)'').  Different sections of the page may have different number of columns. Note that typically such a page does not have any narrative text, e.g.{} attempting to stitch together various pieces of information.

  A key problem for directory pages is determining the {\em reading order}. The typical left-to-right across-the-column / top-to-bottom reading order for English language pages does not work for directory pages since the organizing unit for directory pages is not columns but entries. Consider again Figure~\ref{figure 1a}. A standard reading order would give:

  \begin{quotation}
    ``Legal Counsel to the Fund and the Master Fund'' / ``(as per Hong Kong legal matters)'' / ``(as per Singapore legal matters)'' / ``RAM (LUX) SYSTEMATIC FUNDS 14, boulevard Royal L-2449 LUXEMBOURG'' /  ``BANQUE DE LUXEMBOURG
Société anonyme (public limited company) 14, boulevard Royal L-2449 LUXEMBOURG
''
  \end{quotation}

  But this is incorrect. The reading order is instead defined by the entry. A better reading order would be:
  
  \begin{quotation}
    ``Legal Counsel to the Fund and the Master Fund'' / ``(as per Hong Kong legal matters)'' ``RAM (LUX) SYSTEMATIC FUNDS 14, boulevard Royal L-2449 LUXEMBOURG'' / ``(as per Singapore legal matters)''  ``BANQUE DE LUXEMBOURG Société anonyme (public limited company) 14, boulevard Royal L-2449 LUXEMBOURG''
  \end{quotation}

  Here the reading order for the two entries is de-convolved. But even this is not quite correct. A human scanning the page would conclude that the header ``Legal Counsel to the Fund and the Master Fund'' applies to both the entries below it, so the right structure is:
  
  \begin{quotation}
    ``Legal Counsel to the Fund and the Master Fund''  ``(as per Hong Kong legal matters)'' ``RAM (LUX) SYSTEMATIC FUNDS 14, boulevard Royal L-2449 LUXEMBOURG'' / ``Legal Counsel to the Fund and the Master Fund'' ``(as per Singapore legal matters)''  ``BANQUE DE LUXEMBOURG Société anonyme (public limited company) 14, boulevard Royal L-2449 LUXEMBOURG'' 
  \end{quotation}

  But an even better representation is obtained by recognizing that the text on the page should not be represented as a {\em sequence}, but rather as a {\em tree}, (the {\em reading tree}) as illustrated in Figure~\ref{figure 1b}. Here, the hierarchy captures that the header of a parent node applies also to the child nodes.

 The {\em directory page parsing problem} is thus the problem of recovering the reading tree given the directory page.

More examples of directory pages and their reading trees are provided in Appendix~\label{Appendix-more-examples}.

 \subsection{Our solution}
            
            The key insight behind our solution is that the tree structure can be recovered by a {\em bottom-up parse}, processing the visual elements on the page from the bottom right corner, leftwards and upwards. This corresponds to identifying the leaves of the tree first and then building up the parent nodes. We assume the availability of a library (the {\em visual json} library) that will read the pdf representation of the page and break them up into {\em groups} which consist of {\em lines}, which consist of {\em segments}. Our algorithm, at each stage will maintain a forest of trees (representing a parse of a portion of the page) and the set of segments that have not been processed so far. Initially the tree consists of a single empty (faux) node at the bottom right of the page. The algorithm will choose a segment not in the forest that is closest to some tree in the forest (to the left or above a node in the tree), label it as either a {\tt body} or {\tt header} node, and either extend the neighboring tree (by adding an node on top), or start a new tree.
            To illustrate (Figure~\ref{figure 1a}):
            \begin{enumerate}
            \item The segment ``Oddo Asset Management SA … Cedex 09 France'' is marked as body, and placed in its own tree (Node 1).
            \item The segment ``(as per Cayman Legal Matters)'' is marked as header, placed in a separate entry, and marked the parent of Node 1 (Node 2).
            \item The segment ``BANQUE DE LUXEMBOURG … L-2449 LUXEMBOURG'' is marked as body. Node 2 is deemed not to dominate this body (since Node 2 occurs below it visually), hence a new entry is created (Node 3).
            \item The segment ``RAM (LUX) SYSTEMATIC FUNDS … L-2449 LUXEMBOURG'' is marked as body. This is deemed too far away to be part of the same entry as Node 3, and hence starts a new entry (Node 4).
            \item The segment ``(as per Singapore legal matters)'' is marked as header and seen to continue the entity containing Node 3 (Node 5).
            \item The segment ``(as per Hong Kong legal matters)'' is marked as header and seen to continue the entity containing Node 4 (Node 6).
            \item The segment ``Legal Counsel to the Fund and Master Fund'' is marked as header, and seen to dominate all three entries being built under it, viz. Node 2, Node 5 and Node 6 (Node 7).
            \item The rest of the tree is built in the same way, yielding the reading tree of Figure~ \ref{figure 1b}.
            \end{enumerate}

            Once the reading tree is constructed, the headers and body of each entity can be read off immediately. Entity recognition techniques can be applied to determine organization and (possibly partial) role in the entry. The tree structure is used to complete the role associated with an entry.

\subsection{Related work}
        
A number of techniques have been proposed and used currently to determine the correct Reading Order of text in variety of representations, although all of them have their own set of assumptions. \cite{Clausner} highlights the significance of determining the the correct reading order and rely on OCR engines like Abby and Tesseract to give correct grouping of text for Pages. \cite{Lingcai1}  proposed an approach for article reconstruction using a bi-partite graph framework by making use of the linguistic information to define the correct reading order. This approach proves to be non-relevant for the task at hand, since there is no logical/linguistic continuity or flow of text between blocks in the directory pages. \cite{Ismael} proposed a rule based approach for extracting the correct reading order from PDF documents having multi-column layout in which the text-blocks are handling tree-structured text: parsing directory pages separable by horizontal and vertical cuts. This approach aims to parse the text into a sequential reading order, which could not be applied to directory page parsing because the requirement is to get the hierarchical representation of the text. Moreover, demarcation of text into horizontal and vertical cuts is not always feasible on directory pages as the directory blocks lack sequential ordering. \cite{Smith} used tab stops for deducing the column layout of the page and then assumed that the text inside a column always follows the general reading order. \cite{Phaisarn1} proposed a method to partition the text segments in the documents along the two axes and applied the conventional top to bottom, left to right reading order to the text blocks generated after segmentation. This is an interesting approach but has the drawback of specifying the threshold parameters for the cutting strategy. Parameter like the spacing between the text segment is variant across prospectus documents since they do not come from a common source of origin and hence setting this parameter to a pre-defined optimal value to achieve the correct reading order becomes challenging in our environment. \cite{Stefano} assumed that the header can only corresspond to a single body node which is generally not the case with directory pages.

\subsection{Rest of this paper}
In the rest of the paper we provide details of the algorithm, features used for classification, results and applications of our work in relation extraction and results.

\begin{table}[h!]
  \caption{Hand Crafted Features (w/ importance for Random Forest)}
  \label{tab:Features}
  \begin{tabularx}{\linewidth}{|c|X|c|}
    \toprule
    S.No. & Features & Importance\\
    \midrule 
     1 & \# currency patterns &  \\ \hline
     2 & \# date patterns & \\ \hline
     3 & \# email patterns & \\ \hline
     4 & \# phone numbers patterns  & 0.019\\ \hline
     5 & \# FAC (Buildings, airports, highways, bridges, etc.) entities & \\ \hline
     6 & \# groups & \\ \hline
     7 & Percentage of table on page  & \\ \hline
     8 & \# role entities on page &  \\ \hline
     9 & \# words & 0.212 \\ \hline
     10 & \# groups with Address Candidates  & 0.539\\ \hline 
     11 & \# groups surrounded by borders on all 4 sides  & \\ \hline
     12 & \# groups with \# prospectus orgs $1 \leq  3$  & 0.135\ \\ \hline
     13 & \# groups with roles $1 \leq 4$ & 0.038\\ \hline
     14 & Ratio of \# groups with \# prospectus orgs $1 \leq 3$ to total groups  & \\ \hline
     15 & Ratio of \# groups with roles $1 \leq  4$ to total group &  \\ \hline
    \bottomrule
  \end{tabularx}
\end{table}

\begin{table*}[h!]
  \caption{Top 3 experiment results with Random Forest}
  \label{tab:Random Forest}
\begin{tabularx}{\textwidth}{|c|c|c|c|c|c|c|}

\toprule

\multicolumn{1}{|c|}{Distribution} & \multicolumn{1}{X|}{Features Used} & \multicolumn{4}{c|}{Hyperparameters}  & \multicolumn{1}{X|}{F1 on Train set}\\

\cline{3-6}

\multicolumn{1}{|X|}{} & \multicolumn{1}{X|}{} & \multicolumn{1}{X|}{Maximum Depth Of Trees} & \multicolumn{1}{X|}{Fraction of Features considered while splitting} & \multicolumn{1}{X|}{Minimum \# samples at leaf node} & \multicolumn{1}{X|}{\# trees in the forest} & \multicolumn{1}{X|}{}\\

\midrule

83 pos / 800 neg   & 1-13 &None &0.8 &2 &20 &0.87 \\
83 pos / 800 neg   & 1-13 &20 &0.85 &6 &20 &0.87 \\
83 pos / 800 neg   & 1-13 &10 &0.9 &6&40 &0.85 \\

\bottomrule
\end{tabularx}
\end{table*}

\begin{table*}[h!]
  \caption{Top 3 experiment results with Multi Layer Perceptron}
  \label{tab:MLP}
\begin{tabularx}{\textwidth}{|c|c|c|c|c|c|}

\toprule

\multicolumn{1}{|c|}{Distribution} & \multicolumn{1}{X|}{Features Used} & \multicolumn{1}{c|}{Hidden Layers}& \multicolumn{2}{c|}{Hyperparameters}  & \multicolumn{1}{X|}{F1 on Train set}\\

\cline{4-5}

\multicolumn{1}{|X|}{} & \multicolumn{1}{X|}{} & \multicolumn{1}{X|}{} & \multicolumn{1}{c|}{Epoch} & \multicolumn{1}{c|}{Dropout} & \multicolumn{1}{X|}{} \\

\midrule

166 pos / 2000 neg   & 1-11, 14-15  &One Layer, 128 Neurons, ReLU activation &200 &0.5 &0.81  \\
166 pos / 2000 neg   & 1-11, 14-15  &Two Layer, 256,2 Neurons, ReLU activation &200 &0.25 &0.8  \\
166 pos / 2000 neg   & 1-11, 14-15  &Two Layer, 256,2 Neurons, ReLU activation &200 &0.5&0.8 \\

\bottomrule
\end{tabularx}
\end{table*}

\section{Algorithm Details}

\subsection{Overview}
We use the {\em visualjson} library that takes as input a pdf file and produces a visual json representation, which groups the information in the pdf into {\em pages, groups, lines} and {\em segments}, and identifies {\em page headers} and {\em page footers}.  Below, by {\em style information} we mean information about the appearance of a line, e.g. font style, weight, color and size.

\begin{itemize}
\item A {\em segment} is made up of words which share the same {\em style information}. 
\item A {\em line} is a visual line in the document which is broken down into segments on the basis of {\em style information} or a large amount of space between words. 
\item A {\em group} usually represents sequences of lines that follow a {\em general reading order} and hence constitute a paragraph.  
\item A {\em page} represents a visual page in the document which consists of zero or more groups.
\end{itemize}

We also assume availability of an entity recognizer.

The algorithm has three steps:
\begin{enumerate}

\item Identify directory pages
\item Text segmentation: Each segment on a directory page is tagged as {\tt Header}, {\tt Body} or {\tt Neither}.

\item Reading Tree construction: This step takes a sequence of labeled segments (on a directory page) from the previous step and constructs the reading order tree.
\end{enumerate}

\begin{figure}[h!]

\begin{subfigure}{.5\textwidth}
  \centering
  \includegraphics[scale=0.15]{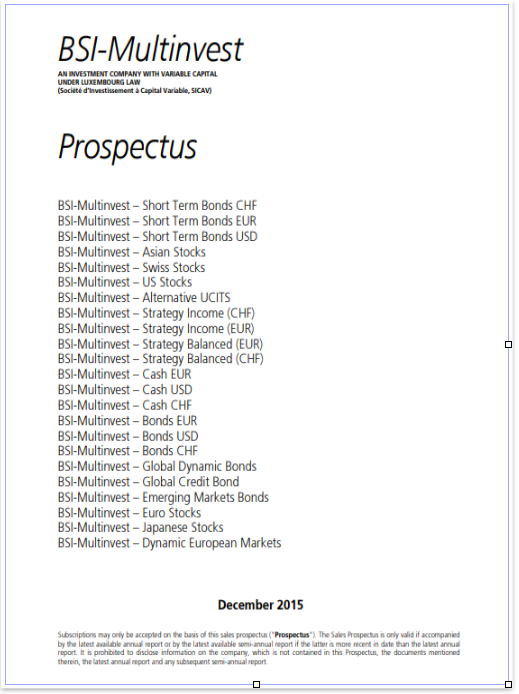}  
  \caption{Org Listing Page}
  \label{fig:Miss 1}
\end{subfigure} 
\begin{subfigure}{.5\textwidth}
  \centering
  \includegraphics[scale=0.15]{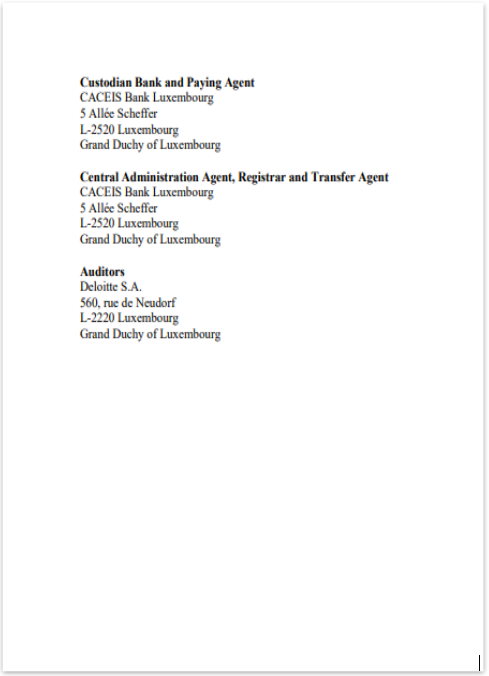}  
  \caption{Directory Page extending beyond 1 page}
  \label{fig:Miss 2}
\end{subfigure} 
\begin{subfigure}{.4\textwidth}
  \centering
  \includegraphics[scale=0.15]{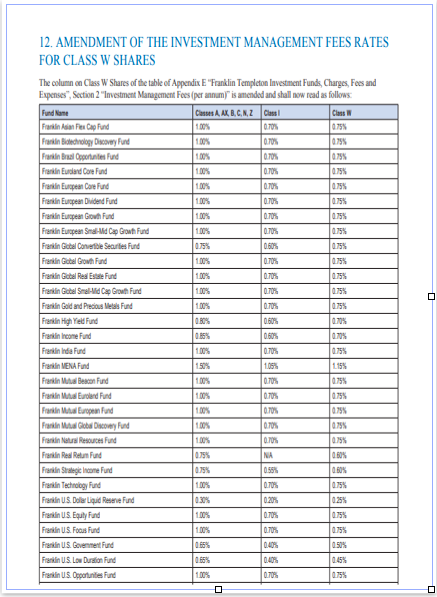}  
  \caption{Page with Tabular Regions}
  \label{fig:Miss 3}
\end{subfigure}

\caption{Misclassified Pages}
\label{fig:Misclassified Pages}
\end{figure}

\subsection{Directory Page Identification}
The task of directory page identification can be modeled as a 2-class exclusive Classification Task. 

The training data consisted of 83 positive samples (directory pages) and 8472 negative samples (non-directory pages), collated from 90 documents. In order to address the high class imbalance, we experimented with resampling the class distribution. The final distributions which resulted in the best performance are given in Table~\ref{tab:Random Forest} and Table~\ref{tab:MLP}.

We experimented with Random Forest and Multi Layer Perceptron models for this Classification task. One of the primary step was to hand craft a set of features that help to distinguish between a {\itshape Directory Page} from a Non-Directory Page.  To begin with, we devised an initial list of 21 features. Feature selection helped in reducing the number of features to 15, listed in Table 1.

Entities (Organizations and Persons) [Feature 12] and their Addresses [Feature 10] give strong evidence of {\itshape Page} being a {\itshape Directory Page}. Since a {\itshape Group} does not contain more than one Address Span, the collective presence of indicators like the GPE, Postcodes and Cardinals is enough to identify the occurrence of an Address. 

After analyzing the results of the initial training iterations, we noticed some common categories of misclassified pages(Figure~\ref{fig:Misclassified Pages}). The pages which were a continuation of a directory page (as shown in Figure~\ref{fig:Miss 2}), but typically with fewer entities were given a negative label. Therefore, features indicating the ratio of the organizations/roles to the number of groups in the page were introduced instead of just the count of such entities [Features 14,15]. Pages like Figure~\ref{fig:Miss 1} which consisted of an organization listing were being falsely marked as directory pages. To overcome this, we introduced a feature indicating the presence of Roles in a page [Feature 8]. Another common false positive page structure had entities listed in form of tables (for e.g. Figure~\ref{fig:Miss 3}). Introduction of features to indicate the presence of tabular structure in the page [Features 7,11] helped in eliminating this False Positive. The final set of features that gave the best performance for both the models are given in Table~\ref{tab:Random Forest} and Table~\ref{tab:MLP}. 

The feature importance for the top 5 features with respect to the Random forest model is given in Table~\ref{tab:Features}.

The test dataset was created using 23 unseen prospectus documents. It had 26 positive samples (directory pages) and 2113 negative samples (non-directory pages). The test results are in Table~\ref{tab:Dir Page Test Results}.

\begin{table}[h!]
  \caption{Performance of Dir Page Classifier on Test data set}
  \label{tab:Dir Page Test Results}
  \begin{tabularx}{\linewidth}{|X|X|X|X|X|}
    \toprule
    Model & Positive-Negative Samples & Precision & Recall &  F-Score\\
    \midrule
    Random Forest & 26 pos / 2113 neg & 0.96 & 0.92 & 0.94\\
    \midrule
    MLP & 26 pos / 2113 neg & 0.8 & 0.92 & 0.86\\
    
    \bottomrule
  \end{tabularx}
\end{table}

\subsection{Directory Page Text Segmentation}

{\itshape Headers} are typically text spans without numbers, they are usually short (2-10 words) and extend over less than 3 Lines. They may have key phrases which indicate the role played by an entity (e.g.{} Investment advisor) or the type of Address (e.g. Registered office). They are generally visually different from the rest of the page.

{\itshape Bodies} are text spans which contain information related to the entity mentioned in its corresponding header. They may have addresses or the name of the entity (e.g. organization or person) playing the role mentioned in the header. They are usually longer than the header, and span  10-30 words.

In order to extract Header and Body Text Spans, we iterate over all the groups of the directory pages and perform the steps mentioned in Algorithm 1.

The results of the segmentation of text are given in the Table~\ref{tab:Text Segmentation Results}

\begin{table}[h!]
  \caption{Results of text Segmentation into Body and Header}
  \label{tab:Text Segmentation Results}
  \begin{tabularx}{\linewidth}{|X|X|X|X|}
    \toprule
    \# Labels & Precision & Recall &  F-Score\\
    \midrule
    1109 & 0.985 & 0.981 & 0.983 \\
    
    \bottomrule
  \end{tabularx}
\end{table}

\begin{algorithm}[h!]
\DontPrintSemicolon
  
   \While{Groups}
   {
    \If{Group represents a Page Footer or Page Header}
    {
        Mark it as Neither
    }

    \If{There is text succeeding the Entity Span in this Group}
    {
        Mark the Span from the start of an Entity to the end of the Group as Body
    }\Else{
        The Entity Span is marked as a Header
    }

    \If{Text of the Group ends with ':' or '-' except for cases like Tel: and Email:}
    {
        Mark it as a Header
    }
    \ElseIf{Group Text contains Roles or Address Types}
    {
      Mark it as a Header
    }
    \Else 
    {
      \If{Remaining Span which  has different color than the predominant color }
      {Mark it as a Header}
      \ElseIf{The text is in Bold or Italic}
      {Mark it as a Header}
      \ElseIf{Font size of the Text is bigger than the majority of the Text in the page}
      {Mark it as a Header}
      \ElseIf{Font Family of the Text is different from the majority of the Text in the Page}
      {Mark it as a Header}
    }

    \If{Any Text Span still remains}
    { Mark it is a Body}
   }

\caption{Text Segmentation into Headers and Bodies}
\end{algorithm}

\subsection{Directory Tree Construction}

The text spans labelled as a {\tt header} or {\tt body} represent the nodes of the directory tree. The root node of the directory tree is a synthetic dummy node. The leaf nodes of the tree are always {\tt body} nodes and all non-leaf nodes are {\tt header} nodes. The path from the root to leaf node provides a natural reading order for the text present in the body nodes. We call this unit of path as a {\tt Directory Block}. The proposed solution leverages a combination of document specific features which depend on the visual structure of the text and generic features which are applicable for determining the reading order on structured pages. An interesting thing to note is, {\itshape Directory Trees} in Figure~\ref{figure 1b} and Figure~\ref{figure 4b} have very similar structure but the visual layouts of the {\itshape Directory pages} from which they are constructed are different. 

\begin{algorithm}[h!]
\DontPrintSemicolon
  
   \For{Each node C in reversed(Nodes)}
   {
    \If{Label(C) == Body}
    {
        \For{ Bi in already traversed Body Nodes}
        {
          \If{parent(Bi) == None and C and Bi have the same immediate parent Header}
          {
            children(C).append(Bi)
          }
          parent(Bi) = C
        }
    }
    \If{Label(C) == Header}
    {
        \For{ Ni in already traversed Nodes}
        {
          \If{parent(Ni) == None}
          {
            \If{Label(Ni) == Header and C can be parent to Header Ni}
            { parent(Ni) = C }
            \If{Label(Ni) == Body and C can be parent to Body Ni}
            { parent(Ni) = C }
          }
        }
    }
   }

\caption{Directory Tree Construction}
\end{algorithm}

\begin{table}[h!]
  \caption{Results of The Directory Tree Construction}
  \label{tab:Dir Tree Construction result}
  \begin{tabularx}{\linewidth}{|X|X|X|X|}
    \toprule
    Evaluation Technique & Precision & Recall &  F-Score\\
    \midrule
    Dir Block & 0.89 & 0.88 & 0.88 \\
    \midrule
    Body and its Immediate Parent & 0.98 & 0.97 & 0.98 \\
    \midrule
    Nodes across Dir Blocks & 0.92 & 0.92 & 0.92 \\
    
    \bottomrule
  \end{tabularx}
\end{table}

The algotihm begins with clustering the {\itshape Header Nodes} together based on their Font Size and then on the basis of visual features like the Font Style, Font Weight, Font Casing and Font Color. Any pair of nodes from the same cluster cannot have a parent child relationship between them because they should be visually distinct from each other. This document specific meta information is used while creating the tree representation.

The Tree representation is built in a bottom-up (from bottom-right, going right to left, bottom to top) manner. The idea behind this bottom up approach is to focus on the regions of the {\itshape Page} which are dominated by the {\itshape Header}. In general, a {\itshape Header} dominates the space below it and to its right, unless the space is already dominated by some other {\itshape Header}. Bottom up traversal, gives one an idea about the dominant space of the {\itshape Headers} and hence helps in establishing the correct links between the {\itshape Body} and {\itshape Header}.  Factors like the Alignment of the text, belongingness to the same {\itshape Group}, adjacency of the nodes in the horizontal and vertical direction and presence of other {\itshape Candidate Headers} directly on top of the current node being examined influences the parent child relationship between nodes. Detailed steps are mentioned in Algorithm 2.

Results for the Directory Tree construction are given in Table~\ref{tab:Dir Tree Construction result}.  Three metrics were used for the evaluation of the tree construction. The {\em First} one evaluates the correctness of the entire Header stack and the Body.  The {\em Second} metric evaluates the correctness of the {\itshape Headers} which are immediately followed by a {\itshape Body} Node and the {\em Last} one evaluates the number of correct Nodes across all {\itshape Directory Blocks}. The {\em Second} metric is the most important for the relation extraction tasks as the {\itshape Header} closest to the {\itshape Body} captures the most specific relation between the {\itshape Header} and the {\itshape Body} Node.

\section{Applications of Directory Page Pipeline}

As per the Directory Tree shown in Figure~\ref{figure 1b}, we will get the following Directory Blocks.

\begin{enumerate}
\item ``DIRECTORY'', ``Registered Office of the Fund'', ``Deutsche Bank (Suisse) S.A. 2'', ``Boulevard Konrad Adenauer, L – 1115 Luxemburg, Grand Duchy of Luxembourg''.

\item ``DIRECTORY'', ``Administrator of the Fund'', ``Deutsche Bank (Suisse) S.A.'', ``4th Floor'', ``Bahnhofquai 9/11, CH-8023 Zurich, Switzerland''

\item ``DIRECTORY'', ``Auditor of the Fund'', ``KPMG Luxembourg Société Coopérative 39, Avenue John F. Kennedy, L–1855 Luxembourg, Grand Duchy of Luxembourg''

\item ``DIRECORY'', ``Legal Counsel to the Fund and Master Fund'', ``(as per Hong Kong Legal Matters)'',``RAM (LUX) SYSTEMATIC FUNDS 14, boulevard Royal L-2449 LUXEMBOURG''

\item ``DIRECTORY'', ``Legal Counsel to the Fund and Master Fund'', ``(as per Singapore Legal Matters)'', ``BANQUE DE LUXEMBOURG Société anonyme (public limited company) 14, boulevard Royal L-2449 LUXEMBOURG''

\item ``DIRECTORY'', ``Legal Counsel to the Fund and Master Fund'', ``(as per Cayman Legal Matters)'', ``Oddo Asset Management SA 12, boulevard de la Madeleine 75440 Paris Cedex 09 France''

\end{enumerate}

Since the reading order has been correctly determined, the entities present in the {\itshape Body} and the {\itshape Headers} can be linked deterministically.  As mentioned in the introduction section, there are a lot of relation extraction tasks which can leverage this reading order.  Now it becomes straightforward to link ``Deutsche Bank (Suisse) S.A.'' to the ``Administrator'' (Directory Block 2) instead of ``Auditor'' which would be the case if we follow the General Reading Order. We can even determine the type of ``Boulevard Konrad Adenauer, L – 1115 Luxemburg, Grand Duchy of Luxembourg.'' as ``Registered Office'' (Directory Block 1) and can identify that ``BANQUE DE LUXEMBOURG'' has jurisdiction only in the Singapore related matters (Directory Block 5)

One of the use-case of the text segmentation output can be Address Span Detection. The nodes that have been identified as Body are more likely to have {\itshape Addresses} in them.  This pre-filtering approach reduces the overall compute time of the Address span detection algorithm, as it has less candidates to process.

\section{Conclusion}

Our approach has achieved high precision/recall numbers in finding the correct Reading Order of the structured content present in {\itshape Directory Pages}. We have demonstrated that using a combination of Document specific and Generic features helps immensely in parsing tree structured text. We have illustrated that the use of {\itshape Directory Blocks} for various relation extraction tasks from {\itshape Directory Pages} boosts their performance significantly. The proposed techniques are generic and can be applied to other {\itshape Document Types} as well having similar kind of tree strucutred text.

\begin{acks}
The authors would like to express their gratitude to Luciano Del Corro and Johannes Hoffart for their valuable and constructive suggestions.
\end{acks}

\bibliographystyle{ACM-Reference-Format}
\bibliography{sample-bib}


\appendix

\label{Appendix-more-examples}

\begin{figure}[b]
\centering
\begin{subfigure}{.5\textwidth}
  \centering
  \includegraphics[width=.6\linewidth]{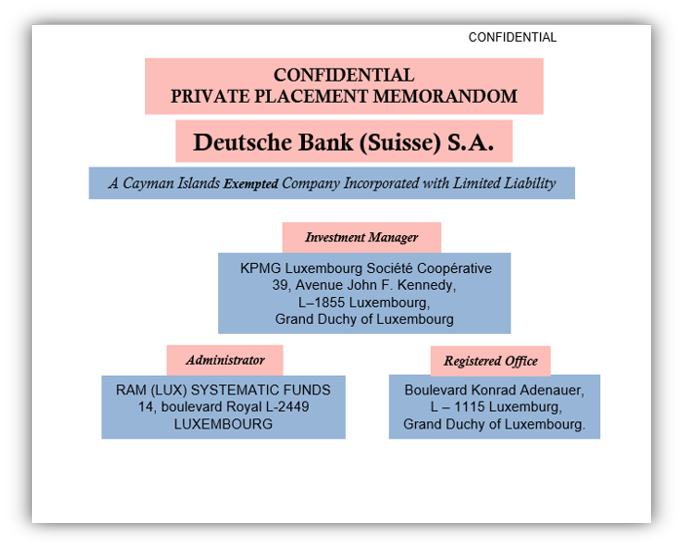}  
  \caption{Example 2}
  \label{fig:Dir Page 1}
\end{subfigure} %
\begin{subfigure}{.5\textwidth}
  \centering
  \includegraphics[width=.8\linewidth]{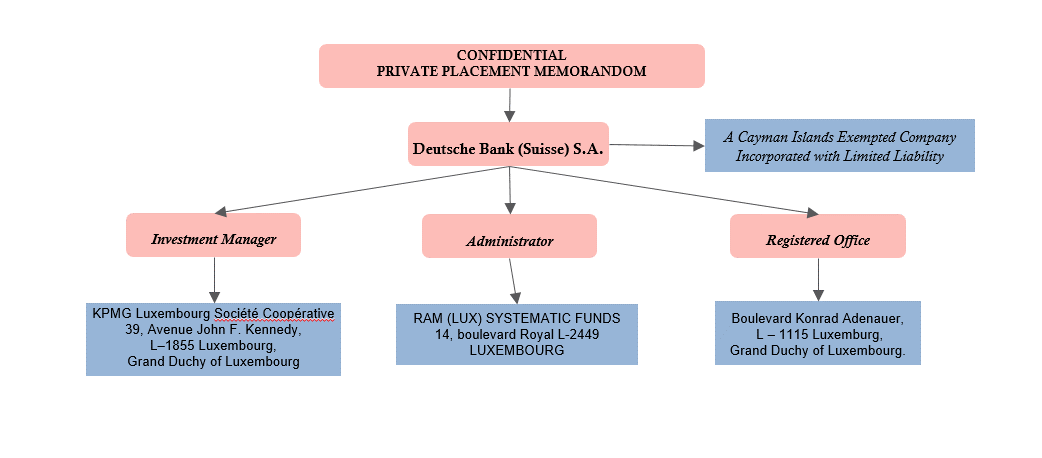}  
  \caption{Directory Tree for Example 2}
  \label{fig:Dir Tree 1}
\end{subfigure}
\begin{subfigure}{.5\textwidth}
  \centering
  \includegraphics[width=.6\linewidth]{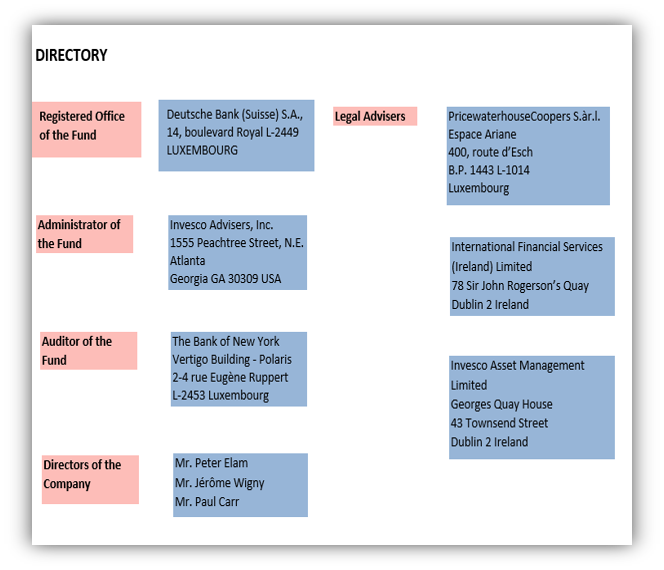}  
  \caption{Example 3}
  \label{fig:Dir Page 2}
\end{subfigure} %
\begin{subfigure}{.5\textwidth}
  \centering
  \includegraphics[width=.8\linewidth]{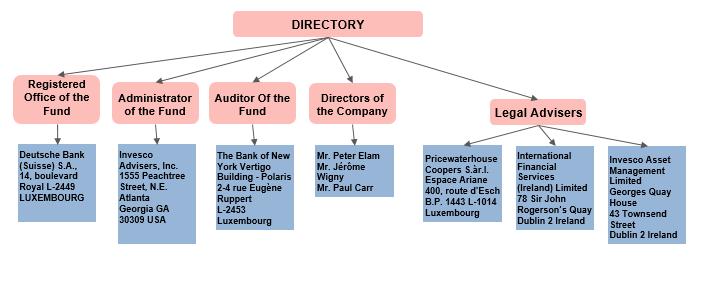}  
  \caption{Directory Tree for Example 3}
  \label{fig:Dir Tree 2}
\end{subfigure}
\caption[Caption]{Directory Pages and their Reading Trees - I\footnotemark[1]}
\label{fig:Dir Trees}
\end{figure}

\begin{figure}[b]
\centering
\begin{subfigure}{.5\textwidth}
  \centering
  \includegraphics[width=.6\linewidth]{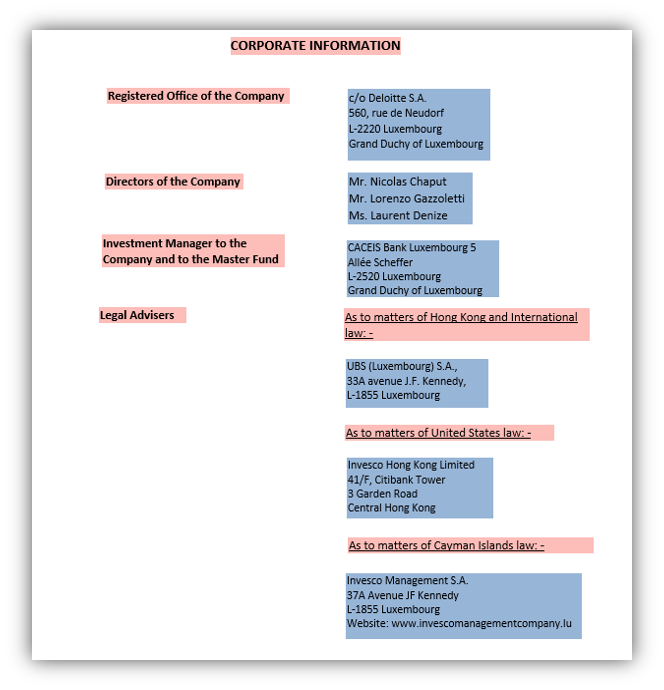}  
  \caption{Example 4}
  \label{figure 4a}
\end{subfigure} %
\begin{subfigure}{.5\textwidth}
  \centering
  \includegraphics[width=.8\linewidth]{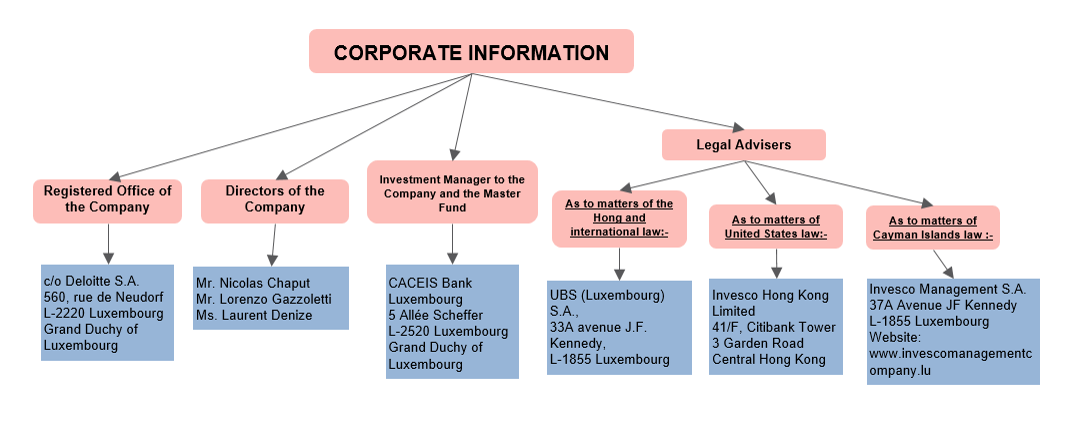}  
  \caption{Directory Tree for Example 4}
  \label{figure 4b}
\end{subfigure}
\begin{subfigure}{.5\textwidth}
  \centering
  \includegraphics[width=.6\linewidth]{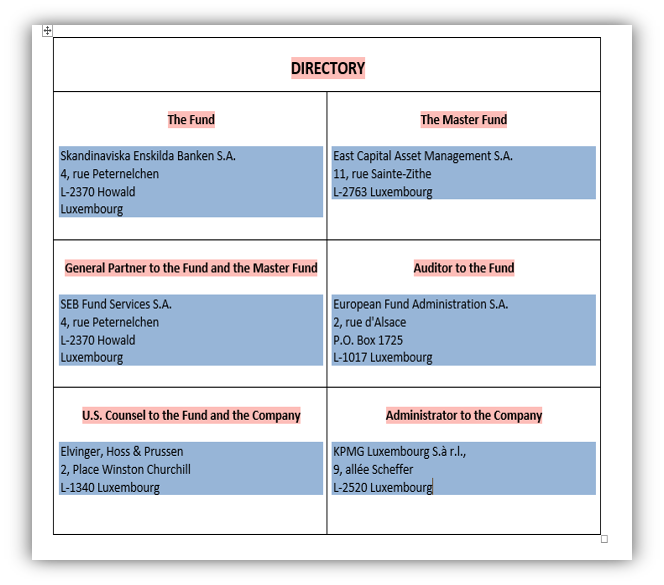}  
  \caption{Example 5}
  \label{figure 5a}
\end{subfigure} %
\begin{subfigure}{.5\textwidth}
  \centering
  \includegraphics[width=.8\linewidth]{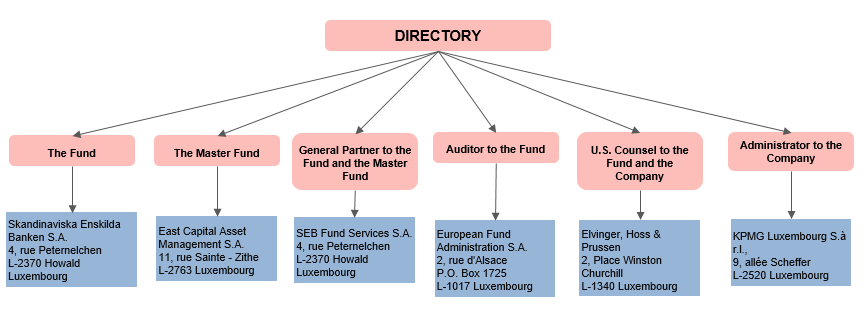}  
  \caption{Directory Tree for Example 5}
  \label{figure 5b}
\end{subfigure}
\caption[Caption]{Directory Pages and their Reading Trees - II\footnotemark[1]}
\label{Dir Trees 3}
\end{figure}

\footnote{Please note that the data present in the Directory Pages and Directory Trees above is synthetically generated}

\end{document}